%% file: root.tex
\documentclass[letterpaper, 10 pt, conference]{ieeeconf}  

\IEEEoverridecommandlockouts                              

\usepackage{graphics} 
\usepackage{graphicx}
\usepackage{float}
\usepackage{verbatim}
\usepackage{mathrsfs}
\usepackage{amsmath}
\usepackage{algorithm}
\usepackage[noend]{algpseudocode}
\usepackage{tabularx} 
\usepackage{booktabs}
\usepackage{multirow}
\usepackage{balance}

\def\etal{\emph{et al.}}

\newcolumntype{L}[1]{>{\raggedright\arraybackslash}p{#1}}
\newcolumntype{C}[1]{>{\centering\arraybackslash}p{#1}}

\newenvironment{myfigure}[1][]{\begin{figure}[#1]\vspace{2mm}}{\end{figure}}
\newenvironment{myfigure*}[1][]{\begin{figure*}[#1]\vspace{2mm}}{\end{figure*}}

\title{\LARGE \bf
Unsupervised Domain Adaptation through Inter-modal Rotation\\for RGB-D Object Recognition$^{*}$}

\author{Mohammad Reza Loghmani$^{1}$, Luca Robbiano$^{2}$, Mirco Planamente$^{2}$, Kiru Park$^{1}$,\\ Barbara Caputo$^{2}$ and Markus Vincze$^{1}$
\thanks{$^{*}$This paper is currently under submission at RA-L with IROS option.}
\thanks{$^{1}$Mohammad Reza Loghmani, Kiru Park and Markus Vincze are with the Vision4Robotics Group, ACIN, TU Wien, Vienna, Austria
        {\tt\small [loghmani, park, vincze]@acin.tuwien.ac.at}}%
\thanks{$^{2}$Luca Robbiano, Mirco Planamente and Barbara Caputo are with the VANDAL Laboratory, Politecnico di Torino and Italian Institute of Technology, Italy
        {\tt\small Luca.Robbiano@studenti.polito.it, [Mirco.Planamente, Barbara.Caputo]@polito.it}}%
}

\begin{document}

\maketitle
\thispagestyle{empty}
\pagestyle{empty}

\begin{abstract}
\input{sections/abstract.tex}
\end{abstract}

\section{INTRODUCTION}
\label{sec:intro}
\input{sections/intro.tex}

\section{RELATED WORK}
\label{sec:related_work}
\input{sections/related.tex}

\section{DATASET}
\label{sec:dataset}
\input{sections/dataset.tex}

\section{METHOD}
\label{sec:method}
\input{sections/method.tex}

\section{EXPERIMENTS}
\label{sec:exp}
\input{sections/expers.tex}

\section{CONCLUSION}
\label{sec:conclusion}
\input{sections/conclusion.tex}
 
\section*{ACKNOWLEDGMENT}
\input{sections/ack.tex}

\balance
\bibliographystyle{IEEEtran}
\bibliography{references}

\end{document}

%% file: sections/abstract.tex
Unsupervised Domain Adaptation (DA) exploits the supervision of a label-rich source dataset to make predictions on an unlabeled target dataset by aligning the two data distributions. In robotics, DA is used to take advantage of automatically generated synthetic data, that come with ``free" annotation, to make effective predictions on real data. However, existing DA methods are not designed to cope with the multi-modal nature of RGB-D data, which are widely used in robotic vision. We propose a novel RGB-D DA method that reduces the synthetic-to-real domain shift by exploiting the inter-modal relation between the RGB and depth image. Our method consists of training a convolutional neural network to solve, in addition to the main recognition task, the pretext task of predicting the relative rotation between the RGB and depth image. To evaluate our method and encourage further research in this area, we define two benchmark datasets for object categorization and instance recognition. With extensive experiments, we show the benefits of leveraging the inter-modal relations for RGB-D DA.
%

%% file: sections/intro.tex
Robotic systems need to recognize objects in order to understand and interact with their surroundings. The basic approach is to perform object recognition on standard RGB images provided by a digital camera. However, the loss of depth information caused by projecting the 3-dimensional world into a 2-dimensional image plane can negatively affect the recognition performance. \mbox{RGB-D} (Kinect-style) cameras provide a potential solution by using range imaging technologies to re-introduce the information about the camera-scene distance as a depth image. While the RGB image contains texture and appearance information, the depth image contains additional geometric information and is more robust to lighting and color variations. The superior information content, coupled with the low sensor price, has caused \mbox{RGB-D} data to be widely used in robot vision.

After the pivotal work of Krizhevsky~\etal~\cite{krizhevsky2012imagenet}, deep convolutional neural networks (CNNs) quickly became the dominant tool in machine vision, establishing new state-of-the-art results for a large variety of tasks, including \mbox{RGB-D} object recognition. The large amount of annotated data required to train CNNs can be very costly and represents one of the main bottlenecks for their deployments in robotics. An attractive workaround that requires no manual annotation consists in generating a large synthetic training set by rendering 3D object models with computer graphics software, such as Blender~\cite{blender}. However, the difference between the synthetic (source) training data and the real (target) test data severely undermines the recognition performance of the network. 
\begin{myfigure}[t]
    \centering
    \includegraphics[width=0.9\linewidth]{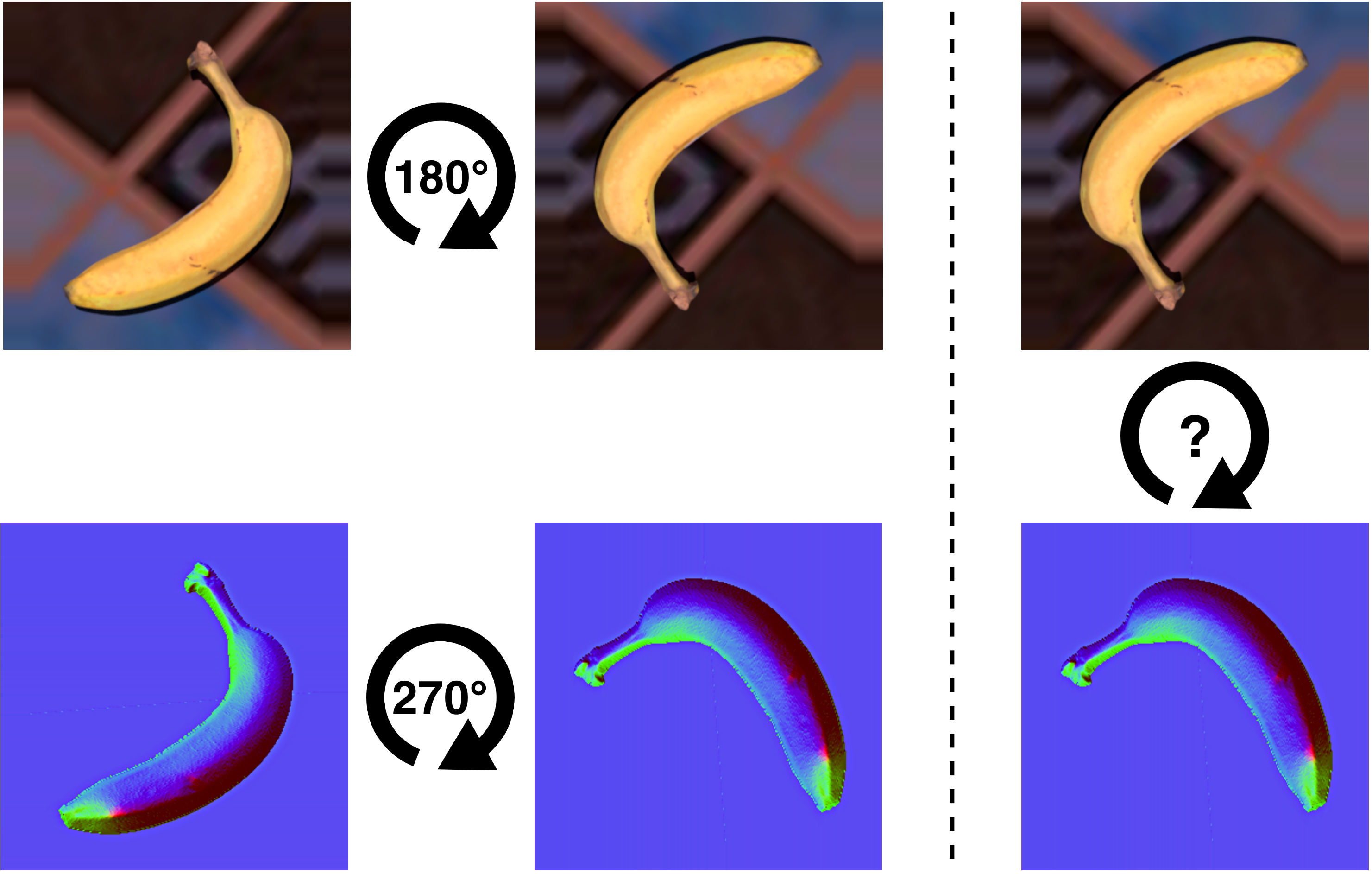}
    \caption{Q: ``By how much should the RGB image (top) be rotated to align with the depth image (bottom)?" A: ``$90^{\circ}$". This question describes the self-supervised task of predicting the relative rotation between the RGB and depth image of a sample after they have been independently rotated. The depth is shown with surface normal colorization~\cite{aakerberg2017improving}.}
    \label{fig:teaser}
\end{myfigure}{}
Unsupervised Domain Adaptation (DA) is a field of research that accounts for the difference between source and target data by considering them as drawn from two different marginal distributions. DA approaches provide predictions on a set of target samples using only annotated source samples, with the unlabeled target samples available transductively. This field has flourished in the last decade and has produced numerous strategies to reduce the shift between the source and target distributions both at feature~\cite{long2015learning, ganin2016domain} and at pixel level~\cite{russo2018from, hoffman2017cycada}. However, existing DA strategies implicitly assume that the data come from a single modality. We claim that this assumption leads to sub-optimal results when dealing with multi-modal data since the natural inter-modal relations of the data are ignored. 

\begin{myfigure*}[t]
    \centering
    \includegraphics[width=0.80\linewidth]{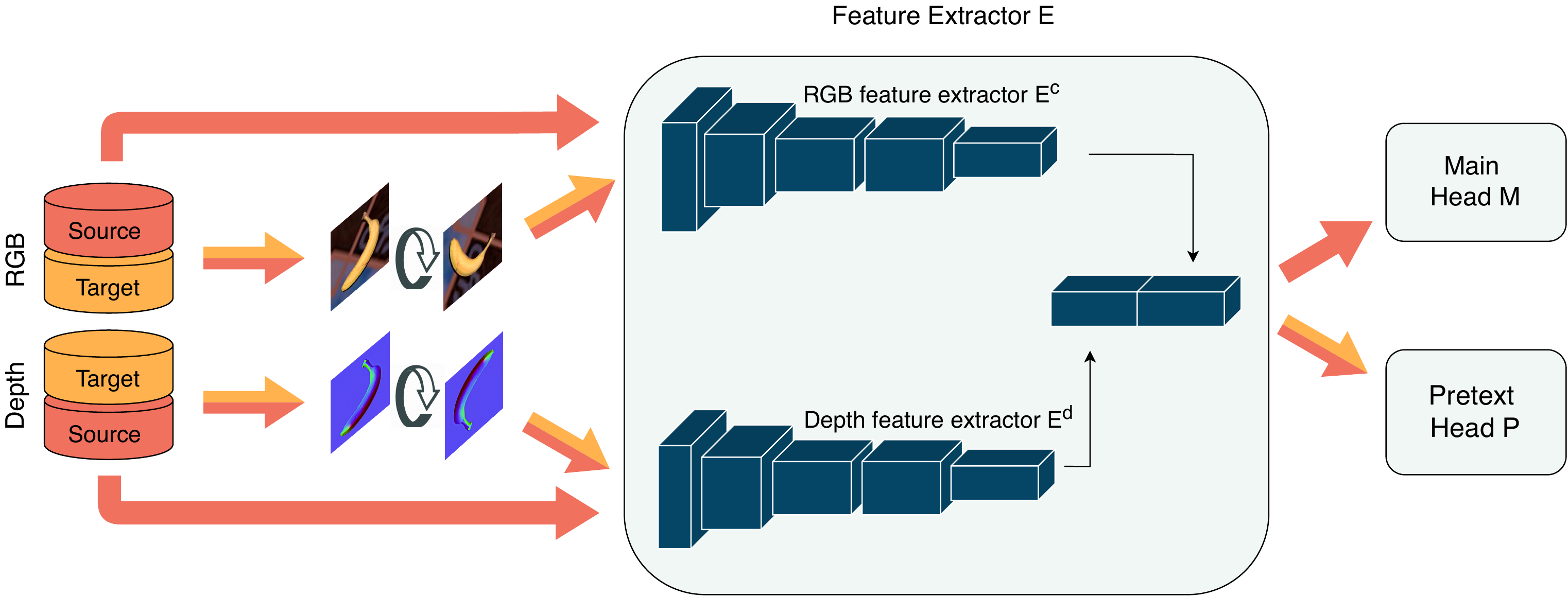}
    \caption{Overview of our method for RGB-D domain adaptation. We use a convolutional neural network (blue squares) that consists of a two-stream feature extractor $E$ that flows into two network heads, the main head $M$ and the pretext head $P$. $M$ is trained for object recognition using the labeled source data (red arrow); $P$ is trained with both source and target samples where the RGB and depth image are independently rotated before being fed to the network (orange+red arrow).}
    \label{fig:architecture}
\end{myfigure*}{}

In this paper, we propose the first DA method tailored to \mbox{RGB-D} data. We define a multi-task learning problem that consists of training a CNN to solve a supervised main task and a self-supervised pretext (or auxiliary) task from pairs of RGB and depth images. The main task is the object recognition problem that we want to solve. The pretext task is an artificial problem created to encourage the network to generate domain-invariant features by learning geometric relations between the RGB and depth modalities: we rotate the RGB and depth image of a sample and ask the network to predict the relative rotation that re-aligns them (see figure~\ref{fig:teaser}). Due to its self-supervised nature, both source and target data can be used to train the model on the pretext task, while the supervision of the source data is used to train the model on the main task (see figure~\ref{fig:architecture}). To evaluate our method on object categorization and instance recognition, we define two benchmark datasets, each composed of a synthetic and a real part. For instance recognition, we render the HomeBrewedDB (HB)~\cite{kaskman2019homebreweddb} models as source dataset and use the real \mbox{RGB-D} sequences of the same dataset as target dataset. For object categorization, no existing dataset presents both synthetic and real data. We use the popular \mbox{RGB-D} Object Dataset (ROD)~\cite{lai2014unsupervised} for the real data and collect the synthetic counterpart ourselves. Therefore, we propose synROD: a dataset generated by collecting and rendering 3D object models from the same categories as ROD using publicly available Web resources. Extensive experiments on these datasets show that our newly defined pretext task effectively reduces the synthetic-to-real domain gap and outperforms existing DA approaches that do not leverage the inter-modal relations of \mbox{RGB-D} data.

In summary, our contributions are the following:
\begin{itemize}
\item a novel multi-modal DA algorithm for \mbox{RGB-D} object recognition that reduces the domain gap by leveraging the relation between RGB and depth data,
\item two benchmark datasets to evaluate RGB-D DA methods on object categorization and instance recognition, including the newly collected synROD, and
\item quantitative and qualitative experiments that showcase the superior performance of our method compared to existing DA approaches.
\end{itemize}

The rest of the paper is organized as follows: the next section positions our approach compared to related work, section~\ref{sec:dataset} describes synROD, section~\ref{sec:method} introduces the proposed method, section~\ref{sec:exp} presents the experimental results and section~\ref{sec:conclusion} draws the conclusions.

%% file: sections/related.tex
\subsection{Unsupervised Domain Adaptation}
The literature of DA can be divided into three groups based on the strategy used to reduce the domain shift. The first group includes discrepancy-based methods~\cite{long2015learning, sun2016return, xu2019larger} that define a metric to measure the distance between source and target data in feature space. The defined metric is then minimized during the training of the network to reduce the domain shift. The second group includes methods based on adversarial learning~\cite{ganin2016domain,tzeng2017adversarial,russo2018from}. In these methods, a domain discriminator and a generator network are trained in an adversarial fashion so that the generator converges to a solution that makes the source and target data indistinguishable for the domain discriminator. The third group includes methods that leverage self-supervised learning to reduce the domain shift~\cite{ghifary2016deep, bousmalis2016domain, xu2019self-supervised, carlucci2019domain}. More precisely, a network is trained to solve an auxiliary self-supervised task on the target (and source) data, in addition to the main task, to learn robust cross-domain representations.

The methods discussed so far are based on the implicit assumption that the data come from a single modality. While no approach specifically aims at adapting multi-modal data when both source and target domain contain RGB-D information\footnote{To our knowledge, Qi~\etal~\cite{qi2018unified} also propose a method to tackle the problem of multi-modal domain adaptation. However, this work does not deal with RGB-D data, but rather focuses on the combination of video and audio and applying it to the RGB-D scenario would require non-trivial adjustments to their method.}, there are a few related cases that are worth mentioning. Spinello and Arras~\cite{spinello2012leveraging}, and Hoffman~\etal~\cite{hoffman2016cross-modal} adapt RGB data to depth data by considering them as source and target domain, respectively. Li~\etal~\cite{li2017domain} tackle the case where the source dataset is composed of RGB-D images, while the target dataset only contains RGB images. The focus is therefore on how to combine the RGB and depth data of the source domain rather than on the adaptation itself. Finally, Wang and Zhang~\cite{jing2019unsupervised} consider the case where both source and target data are RGB-D images, but ignore the multi-modal nature of the data and apply a standard domain adversarial method to reduce the domain shift. In section~\ref{sec:exp}, we provide empirical evidence that this solution is sub-optimal and better results can be achieved by leveraging the relation between the RGB and depth modalities.
\subsection{Self-Supervised Visual Tasks}
Self-supervised learning is used to compensate for the lack of annotated data by training the network on a pretext task for which the supervision (or ground truth) can be defined from the data themselves. Several self-supervised tasks have been defined to tackle computer vision tasks. Some examples include predicting the location of a patch~\cite{doersch2015unsupervised}, solving a jigsaw puzzle~\cite{noroozi2016unsupervised}, colorizing a gray-scale image~\cite{zhang2016colorful}, and inpainting a removed patch~\cite{pathak2016context}. Arguably, one of the most effective self-supervised tasks consists of rotating the input images by multiples of $90^{\circ}$ and training the network to predict the rotation of each image~\cite{gidaris2018unsupervised}. This pretext task has been successfully used in a variety of applications such as network pre-training~\cite{gidaris2018unsupervised}, anomaly detection~\cite{golan2018deep}, and domain adaptation~\cite{xu2019self-supervised}. Inspired by this success, we revisit the rotation prediction task for RGB-D data and propose a novel task that consists of predicting the relative rotation between the color and depth image.


%% file: sections/dataset.tex
In this section, we present synROD and the protocol followed for its creation. More specifically, section \ref{subsec:selection} describes the criteria used to define the scope of the dataset and collect the 3D object models from Web resources; section \ref{subsec:rendering} illustrates the procedure used to render 2.5D scenes from the 3D object models. The dataset will be publicly available upon acceptance of the paper.
\subsection{Selecting 3D Object Models}
\label{subsec:selection}
RGB-D DA has not been explored in the literature yet, so there are no standard benchmark datasets to evaluate methods developed for this purpose. The main challenge of defining a dataset to evaluate DA methods is to identify two distinct sets of data that exhibit the same annotated classes but have been collected in different conditions. In particular, we are interested in the synthetic-to-real domain shift, where the source domain presents RGB-D synthetic data, while the target domain presents RGB-D real data. Existing 3D object datasets, such as ModelNet~\cite{wu20153d} and ShapeNet~\cite{yi2017large-scale}, do not have a corresponding real dataset that shares the same classes. In addition, the lack of texture for some models makes them unusable for our purpose where we are interested in both the shape (depth) and the texture (color) of the object. To overcome this problem, we collect a new synthetic dataset called synROD. We selected the object models for synROD in such a way that each one belongs to one of the 51 categories defined by ROD, arguably the most used RGB-D dataset in robotics for object categorization~\cite{loghmani2019recurrent,carlucci2018deco,eitel2015multimodal,aakerberg2017improving}. We query the objects from the free catalogs of public 3D model repositories, such as 3D Warehouse and Sketchfab, and only keep models that present texture information to be able to render the RGB modality in addition to the depth. All models are processed to harmonize the scale and canonical pose prior to the rendering stage. The final result of the selection stage is a set of $303$ textured 3D models from the 51 object categories of ROD, for an average of about $6$ models per category.
 \begin{myfigure*}[t]
      \centering
      \includegraphics[width=0.90\linewidth]{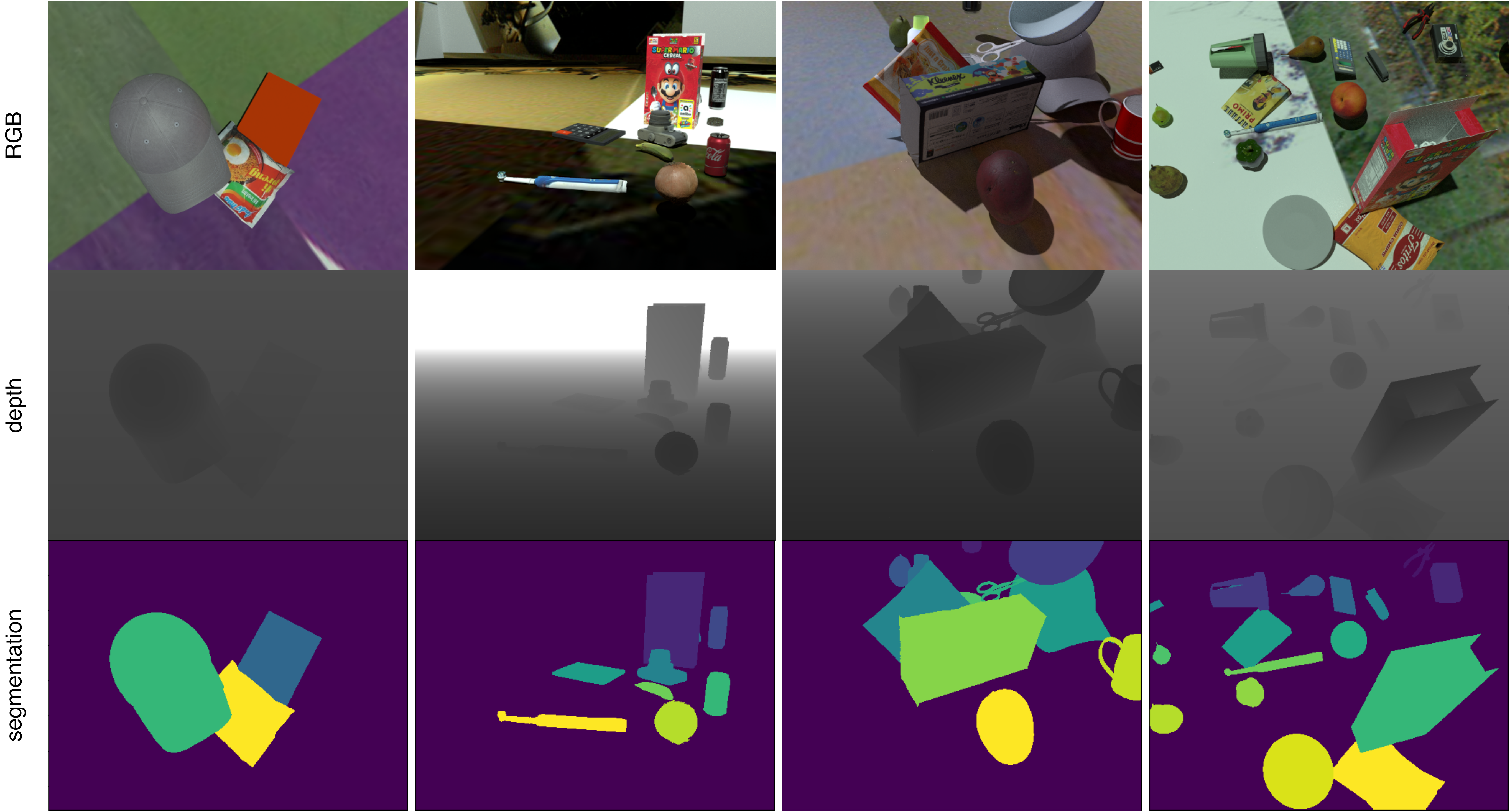}
      \caption{Examples of rendered scenes from synROD with increasing level of clutter from left to right. For each, we showcase the RGB, raw depth and segmentation mask image.}
      \label{fig:scenes_synROD}
  \end{myfigure*}
\subsection{Rendering 2.5D scenes}
\label{subsec:rendering}
We render 2.5D scenes using a ray-tracing engine in Blender to simulate photorealistic lighting. Each scene consists of a rendered view of a randomly selected subset of the models placed on a $1.2 \times 1.2$ meter virtual plane. The poses of the camera and the light source are sampled from an upper hemisphere of the plane with varying radius. To obtain natural and realistic object poses, each model is dropped on the virtual plane using a physics simulator. The number of objects in each scene varies from five to $20$ to create different levels of clutter. To ensure a balanced dataset, we condition the selection of the models to insert in every scene to the number of past appearances. The background of the virtual space containing the objects is randomized by using images from the MS-COCO dataset~\cite{lin2014microsoft}. We rendered approximately $30,000$ RGB-D scenes with semantic annotation at pixel level (see figure~\ref{fig:scenes_synROD}).

%% file: sections/method.tex
In this section, we present our method for RGB-D DA. More specifically, section \ref{subsec:overview} provides a high-level overview of the method, section \ref{subsec:pretext} describes the details of the relative rotation task, section \ref{subsec:architecture} and \ref{subsec:optimization} specify the architecture and training/test protocol of the CNN.
\subsection{Overview}
\label{subsec:overview}
Our goal is to train a neural network to predict the object class of the target data, using only labeled source data and unlabelled target data. We formulate our problem as a multi-task classification by training the network to solve a main supervised task and a pretext self-supervised task. The main task consists of using the supervision of the source data to learn to predict object labels. The pretext task consists of predicting the relative rotation between a pair of RGB and depth images that have been independently rotated. Since the ground truth for this simple pretext task can be generated automatically from the data, we can train the network to predict the relative rotation using both source and target data in a self-supervised fashion. Learning this inter-modal relation yields domain-invariant features and consequently improves the object class prediction on the target data without the need for direct supervision.

\subsection{Pretext Task}
\label{subsec:pretext}
Predicting image rotation is a simple yet effective pretext task to learn robust visual representations~\cite{gidaris2018unsupervised, golan2018deep, xu2019self-supervised}. This self-supervised task consists in rotating a given image by a multiple of $90^{\circ}$ and training a CNN to predict the rotation that has been applied. However, predicting the rotation of an individual image is only possible with datasets such as PACS~\cite{li2017deeper} where the pose of the subject is coherent throughout the samples. For example, the giraffe images in PACS always represent the animal in an upright position. For datasets where the object appears in a variety of poses, predicting the image rotation is an ill-posed problem (see figure~\ref{fig:absolute_rotation}). To overcome this issue and adapt the task to RGB-D data, we define the task of predicting the relative rotation between the RGB image $x^c$ and depth image $x^d$ of an RGB-D sample. Let us denote with $rot90(x,i), i \in [0,3]$ the function that rotates clockwise a 2D image $x$ by $i*90^{\circ}$. Given an RGB-D sample $(x^c, x^d)$, we select $j, k \in [0,3]$ at random to compute $\tilde{x}^c=rot90(x^c,j)$ and $\tilde{x}^d=rot90(x^d,k)$, and indicate with $z$ the one-hot encoded label indicating the relative rotation between them. More precisely, the relative rotation label is computed as $z=one\_hot((k-j)\ mod\ 4)$, where \textit{one\_hot(.)} is the function that generates the one-hot encoding and \textit{mod} is the modulo operator. The pretext task consists of predicting $z$ given $(\tilde{x}^c, \tilde{x}^d)$, or in other words: ``how many times should the RGB image be rotated by $90^{\circ}$ clockwise to align with the depth image?".  Figure~\ref{fig:rotation_examples} shows all the possible combinations for which a pair of RGB and depth images can be rotated and their corresponding relative rotation.
\begin{myfigure}[t]
    \centering
    \includegraphics[width=0.95\linewidth]{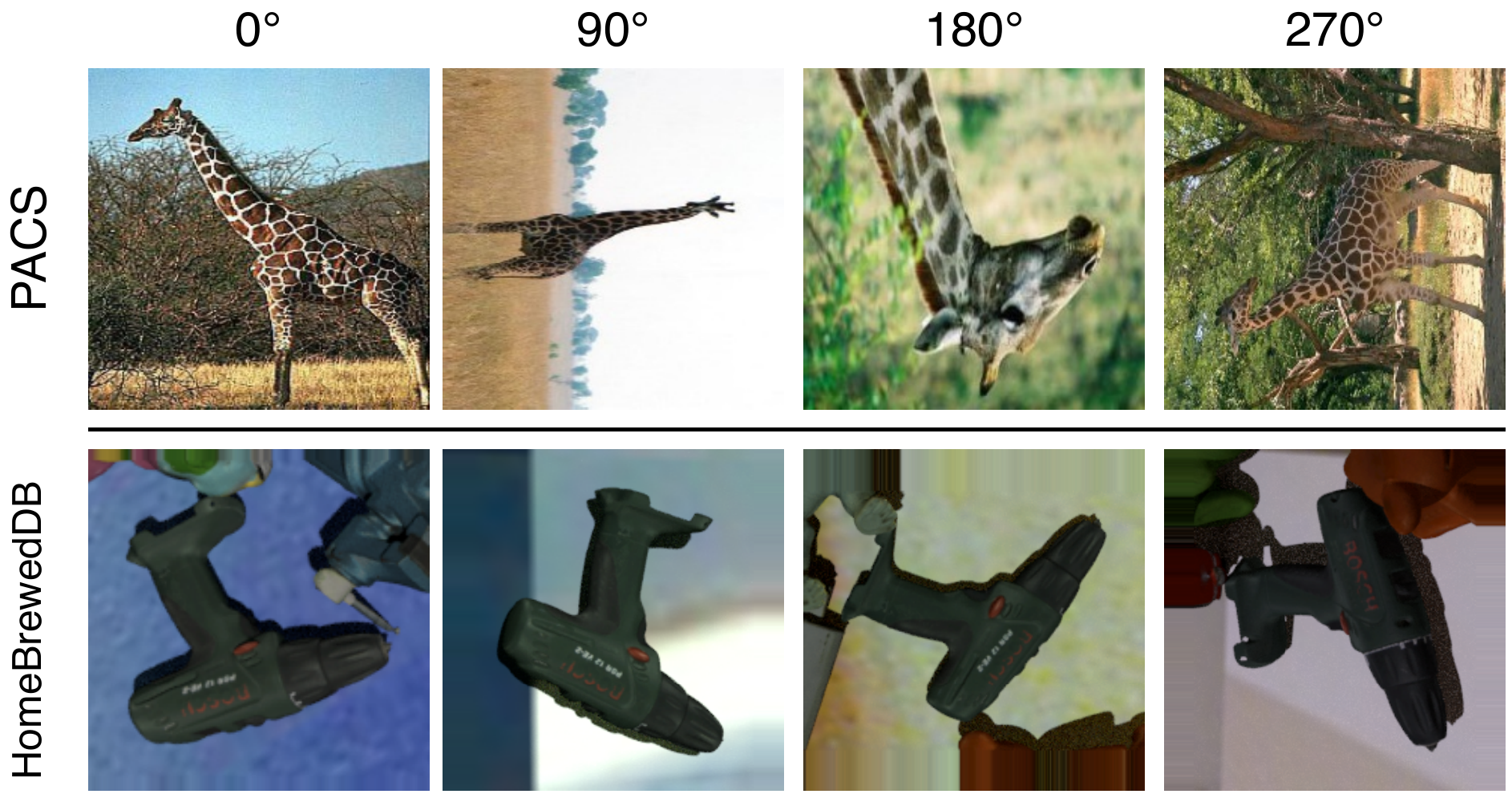}
    \caption{Examples images from PACS~\cite{li2017deeper} (top row) and HomebrewedDB~\cite{kaskman2019homebreweddb} (bottom row) that are rotated by $0^{\circ}$, $90^{\circ}$, $180^{\circ}$, and $270^{\circ}$. It is easy to guess the rotation of the PACS samples based on the background and our prior knowledge of the subject, while the same does not hold for the HomebrewedDB samples. This illustrates why predicting the image rotation by looking at each image individually, as in~\cite{gidaris2018unsupervised}, is an ill-posed task.}
    \label{fig:absolute_rotation}
\end{myfigure}{}
\begin{myfigure}[t]
    \centering
    \includegraphics[width=0.95\linewidth]{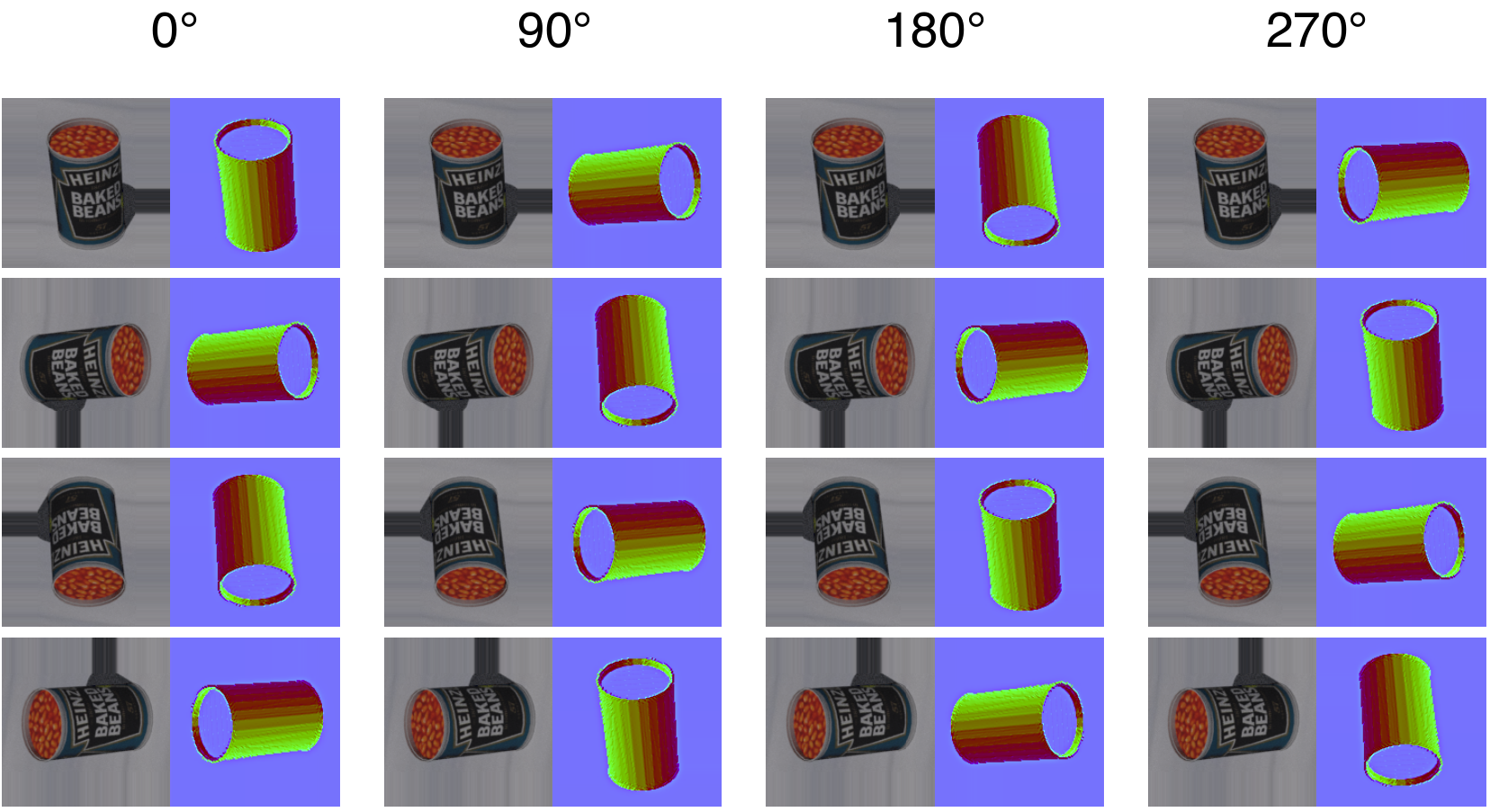}
    \caption{All the possible combinations of RGB and depth rotation for a given relative rotation $\{0^{\circ}, 90^{\circ}, 180^{\circ}, 270^{\circ}\}$.}
    \label{fig:rotation_examples}
\end{myfigure}{}
%
%
\begin{algorithm}
\caption{RGB-D Domain Adaptation}
\label{alg:rgbd_da}
\begin{algorithmic}
\Require 
\State Labeled source dataset $S=\{((x^{sc}_{i},x^{sd}_{i}),y^s_{i})\}_{i=1}^{N_s}$
\State Unlabeled target dataset $T=\{((x^{tc}_{i},x^{td}_{i})\}_{i=1}^{N_t}$
\Ensure
\State Object class prediction for the target data $\{\hat{y}^t_{i}\}_{i=1}^{N_t}$
\State 
\Procedure{training}{S,T}
\State Get transformed set $\widetilde{S}=\{((\tilde{x}^{sc}_{i},\tilde{x}^{sd}_{i}),z^s_{i})\}_{i=1}^{\widetilde{N}_s}$
\State Get transformed set $\widetilde{T}=\{((\tilde{x}^{tc}_{i},\tilde{x}^{td}_{i}),z^t_{i})\}_{i=1}^{\widetilde{N}_t}$
\For{\textbf{each} iteration}
\State Load mini-batch from S
\State Compute main loss $\mathcal{L}_{m}$
\State Load mini-batches from $\widetilde{S}$ and $\widetilde{T}$
\State Compute pretext loss $\mathcal{L}_{p}$
\State Update weights of M from $\nabla \mathcal{L}_{m}$
\State Update weights of P from $\nabla \mathcal{L}_{p}$
\State Update weights of E from $\nabla \mathcal{L}_{m}$ and $\nabla \mathcal{L}_{p}$
\EndFor
\EndProcedure
\Procedure{test}{T}
\For{\textbf{each} $(x^{tc}_{i},x^{td}_{i})$ in $T$}
\State Compute $\hat{y}^t_{i}=M(E(x^{ct}_{i},x^{dt}_{i}))$
\EndFor
\EndProcedure
\end{algorithmic}
\end{algorithm}
\subsection{Network architecture}
\label{subsec:architecture}
Figure~\ref{fig:architecture} shows the structure of the CNN we use for our method. A feature extractor $E$ generates RGB-D features that are provided as input to both the main head $M$ and the pretext head $P$. Each of these modules is a neural network defined with differentiable operation, so the whole network can be trained end-to-end using standard backpropagation.

\textit{Feature extractor:} Following the literature of RGB-D object recognition~\cite{eitel2015multimodal, aakerberg2017improving, carlucci2018deco}, we use a two-stream CNN with a late fusion approach to generate RGB-D features. In other words, two identical CNNs, $E^c$ and $E^d$, are used to process the RGB and depth image, respectively. The outputs of these two networks are then concatenated along the channel dimension to compose the final RGB-D feature. For our experiments, we define $E^c$ and $E^d$ as the ResNet-18~\cite{he2016deep} architecture without the final fully connected and global average pooling layers.

\textit{Main head:} The network $M$ solves a $\mathcal{C}$-way classification problem, where $\mathcal{C}$ indicates the number of object classes we want to predict. It is defined as \textit{[gap, fc(1000), fc($\mathcal{C}$)]}, where \textit{gap} indicates a global average pooling operation, \textit{fc(n)} indicates a fully connected layer with $n$ neurons. \textit{fc(1000)} uses batch normalization and ReLU activation function, while \textit{fc($\mathcal{C}$)} uses softmax activation function.

\textit{Pretext head:} The network $P$ solves the $4$-way classification problem of predicting the rotation between the RGB and depth image. It is defined as \textit{[conv(1 $\times$ 1,100), conv(3 $\times$ 3, 100), fc(100), fc(4)]}, where \textit{conv(k $\times$ k, n)} indicates a 2D convolutional layer with kernel size $k \times k$ and $n$ neurons. All convolutional and fully connected layers use batch normalization and ReLU activation function, except for \textit{fc(4)} that uses softmax activation function. It is worth mentioning that, differently from $M$, we use convolutional layers in $P$ to better preserve the spatial information. In section~\ref{subsec:results} we show that this leads to superior performance compared to adopting the architecture of $M$ for both heads. 

\subsection{Optimization}
\label{subsec:optimization}
Let us denote with $S=\{((x^{sc}_{i},x^{sd}_{i}),y^s_{i})\}_{i=1}^{N_s}$ the set of labeled source data and $T=\{((x^{tc}_{i},x^{td}_{i})\}_{i=1}^{N_t}$ the set of unlabeled target data, where $(x^{*c},x^{*d})$ denotes the pair of RGB and depth images of a sample and $y^s$ denotes the one-hot encoded object class label. From $S$ and $T$, we can generate a transformed set of source and target data, $\widetilde{S}=\{((\tilde{x}^{sc}_{i},\tilde{x}^{sd}_{i}),z^s_{i})\}_{i=1}^{\widetilde{N}_s}$ and $\widetilde{T}=\{((\tilde{x}^{tc}_{i},\tilde{x}^{td}_{i}),z^t_{i})\}_{i=1}^{\widetilde{N}_t}$, that is used to define the relative rotation task. We train the CNN to minimize the objective function $\mathcal{L}=\mathcal{L}_m(y^s,\hat{y}^s)+ \lambda_p \mathcal{L}_p(z^s,\hat{z}^s,z^t,\hat{z}^t)$, where $\mathcal{L}_m$ and $\mathcal{L}_p$ are respectively the cross-entropy loss of the main and pretext task, and $\lambda_p$ is a weight to regulate the contribution of the corresponding pretext loss term. More precisely
\begin{align}
    &\mathcal{L}_m=-\frac{1}{N_s}\sum_{i=1}^{N_s}y^s_i\cdot \log(\hat{y}^s_i),\\
    &\mathcal{L}_p =-\frac{1}{\widetilde{N}_s}\sum_{i=1}^{\widetilde{N}_s}z^s_i\cdot \log(\hat{z}^s_i) -\frac{1}{\widetilde{N}_t}\sum_{j=1}^{\widetilde{N}_t}z^t_j\cdot \log(\hat{z}^t_j),
\end{align}
where $\hat{y}^s=M(E(x^{sc},x^{sd}))$ and $\hat{z}^*=P(E(\tilde{x}^{*c},\tilde{x}^{*d}))$.
At test time, the pretext head $P$ is discarded and the predictions of the target data are computed as $\hat{y}^t=M(E(x^{ct},x^{dt}))$. The pseudo-code is presented in algorithm~\ref{alg:rgbd_da}.

%

%% file: sections/expers.tex
In this section, we present the experimental protocol and the evaluation results of our method. More precisely, section~\ref{subsec:datasets} describes the adopted datasets, section~\ref{subsec:baselines} presents the baseline methods we compare against our method, section~\ref{subsec:implementation} presents the implementation details for training the CNN, and section~\ref{subsec:results} show quantitative and qualitative results on RGB-D DA.
\subsection{Datasets}
\label{subsec:datasets}
\textit{ROD \& synROD:} Since its release in 2011, ROD has become the main reference dataset for RGB-D object recognition in the robotics community. It contains 41,877 RGB-D images of 300 objects commonly found in house and office environments grouped in 51 categories. Each object is recorded on a turn-table with the RGB-D camera placed at approximately one meter distance at $30^{\circ}$, $45^{\circ}$ and $60^{\circ}$ angle above the horizon. As mentioned in section~\ref{sec:dataset}, synROD is a synthetic dataset created using object models from the same categories as ROD.
To make the two datasets comparable, we randomly select and extract approximately 40,000 objects crops from synROD to match the dimensions of ROD. In our experiments, we evaluate RGB-D DA methods by considering synROD as the synthetic source dataset and ROD as the real target dataset.

\textit{HomebrewedDB:} It is a more recent dataset used for 6D pose estimation that features 17 toy,
8 household and 8 industry-relevant objects, for a total of 33 instances. HB provides high-quality object models reconstructed using a 3D scanner and 13 validation sequences. Each sequence contains three to eight objects on a large turntable and is recorded using two RGB-D cameras at $30^{\circ}$ and $45^{\circ}$ angle above the horizon. To re-purpose this dataset for the instance recognition problem, we extract the object crops from all the validation sequences, for a total of 22,935 RGB-D samples, and we refer to it as realHB. We create a synthetic version of this dataset by rendering the reconstructed object models using the same procedure used for synROD (see section~\ref{sec:dataset}), and we refer to it as synHB. In order to make the two datasets comparable, we randomly select and extract about $25,000$ objects crops from synHB to match the dimensions of realHB. In our experiments, we evaluate RGB-D DA methods by considering synHB as the synthetic source dataset and realHB as the real target dataset.
\subsection{Baseline methods}
\label{subsec:baselines}
For our baselines, we consider four different DA methods: \textit{MMD}~\cite{long2015learning}, \textit{GRL}~\cite{ganin2016domain}, \textit{Rotation}~\cite{xu2019self-supervised} and \textit{AFN}~\cite{xu2019larger}. The first two are arguably the most used and well-established DA methods; \textit{AFN} is chosen as the current state of the art, while \textit{Rotation} is the most relevant to our method.

\textit{MMD:} Long \etal~\cite{long2015learning} encourage the final layers of a neural network to generate domain-invariant features by minimizing the empirical maximum mean discrepancy, a metric that measures the discrepancy between two domain distributions.

\textit{GRL:} Ganin \etal~\cite{ganin2016domain} encourage the feature extractor to generate domain-invariant features using adversarial learning. A domain discriminator is trained to distinguish source from target samples, while the feature extractor is trained to fool the discriminator using a gradient reversal layer.

\textit{AFN:} Xu \etal~\cite{xu2019larger} observe that, in the absence of an explicit adaptation, the target data present a significantly lower average feature norm that the source data. Therefore, the feature norm of the one-to-last layer of the network is iteratively increased for both domains to achieve adaptation.

\textit{Rotation:} Xu \etal~\cite{xu2019self-supervised} encourage the feature extractor to generate domain-invariant features by predicting the absolute rotation~\cite{gidaris2018unsupervised} of an RGB image as a pretext task. 

Since the aforementioned methods are not originally designed for multi-modal data, we use two strategies to evaluate their performance on RGB-D DA. First, we adapt each modality separately until convergence and then we freeze the feature extractors and train a fully connected layer on the concatenation of the adapted features (RGB-D). Second, similarly to our method, we apply them to the concatenation of the RGB and depth features generated by the feature extractor $E$,  and train the network in an end-to-end fashion (RGB-D e2e). Finally, we also report the results on the single modalities to verify if it is beneficial to use multi-modal data.
\subsection{Implementation details}
\label{subsec:implementation}
The CNN is trained using SGD optimizer with momentum $0.9$, learning rate $3 \times 10^{-4}$, batch size $64$. Following~\cite{xu2019self-supervised, xu2019larger,morerio2018minimal-entropy}, we include entropy-minimization with weight $0.1$ as a DA-specific regularization, in addition to the more general weight decay $0.05$ and dropout $0.5$. The weights of the two ResNet-18, $E^c$ and $E^d$, are initialized with values obtained by pre-training the networks on ImageNet~\cite{deng2009imagenet}, while the rest of the network is initialized with Xavier initialization. All the parameters of the network, including the pre-trained parameters, are updated during training. The input to the network is synchronized RGB and depth images pre-processed following the procedure in~\cite{aakerberg2017improving}, where the depth information is colorized with surface normal encoding. This technique prevails as the best non-learned depth colorization method to effectively exploit networks pre-trained on ImageNet~\cite{carlucci2018deco} and is widely adopted by state-of-the-art methods for RGB-D object recognition~\cite{loghmani2019recurrent}. 
%
\setlength{\tabcolsep}{7pt}
\begin{table}[t]
\begin{center}
\vspace{3mm}
\caption{Accuracy (\%) of several methods for RGB-D domain adaptation on two synthetic-to-real shifts, synROD$\rightarrow$ROD and synHB$\rightarrow$realHB. Bold: highest result.}
\label{tab:da}
\begin{tabular}{llcc}
\toprule\noalign{\smallskip}
\multicolumn{4}{c}{\normalsize\textsc{RGB-D Domain Adaptation}} \\
\noalign{\smallskip}
\cline{1-4}
\noalign{\smallskip}
Method & & synROD$\rightarrow$ROD & synHB$\rightarrow$realHB\\
\noalign{\smallskip}
\toprule\noalign{\smallskip}
\multirow{4}{*}{\centering Source only} & RGB &
 52.13 & 51.17 \\
& depth &
  7.56 & 15.50 \\
 & RGB-D &
  50.57 & 49.71 \\
 & RGB-D e2e &
  47.70 & 49.45 \\
 \hline
 \noalign{\smallskip}
\multirow{4}{*}{\centering GRL~\cite{ganin2016domain}} & RGB &
 57.12 & 74.74 \\
& depth &
  26.11 & 29.52 \\
 & RGB-D &
  59.09 & 75.23 \\
 & RGB-D e2e &
  59.51 & 74.95 \\
 \hline
 \noalign{\smallskip}
\multirow{4}{*}{\centering MMD~\cite{long2015learning}} & RGB &
 63.68 & 74.95 \\
& depth &
  29.34 & 28.24 \\
 & RGB-D &
  62.10 & 77.96 \\
 & RGB-D e2e &
  62.57 & 77.26 \\
 \hline
 \noalign{\smallskip}
\multirow{4}{*}{\centering Rotation~\cite{xu2019self-supervised}} & RGB &
 63.21 & 84.46 \\
& depth &
  6.70 & 5.62 \\
 & RGB-D &
  63.33 & 83.99 \\
 & RGB-D e2e &
  57.89 & 84.15 \\
 \hline
 \noalign{\smallskip}
\multirow{4}{*}{\centering AFN~\cite{xu2019larger}} & RGB &
 64.63 & 84.04 \\
& depth &
  30.72 & 31.67 \\
 & RGB-D &
  61.19 & 83.06 \\
 & RGB-D e2e &
  62.40 & 86.49 \\
 \hline
 \noalign{\smallskip}
 \textbf{Ours} & & 
 \textbf{66.68} & \textbf{87.28} \\
\noalign{\smallskip}
\bottomrule
\end{tabular}
\end{center}
\end{table}
\setlength{\tabcolsep}{1.4pt}
\setlength{\tabcolsep}{7pt}
\begin{table}[t]
\begin{center}
\caption{Accuracy (\%) of variations of our method for RGB-D domain adaptation on two synthetic-to-real shifts, synROD$\rightarrow$ROD and synHB$\rightarrow$realHB. Bold: highest result.}
\label{tab:ablation}
\begin{tabular}{lccc}
\toprule\noalign{\smallskip}
\multicolumn{4}{c}{\normalsize\textsc{Ablation Study}} \\
\noalign{\smallskip}
\cline{1-4}
\noalign{\smallskip}
Method & synROD$\rightarrow$ROD & synHB$\rightarrow$realHB & avg. drop\\
\noalign{\smallskip}
\toprule\noalign{\smallskip}
Target rotation & 63.60 & 86.32 & 2.03\\
FC classifier & 64.20 & 86.49 & 1.64\\
 \hline
 \noalign{\smallskip}
 \textbf{Ours} & \textbf{66.68} & \textbf{87.28} & - \\
\noalign{\smallskip}
\bottomrule
\end{tabular}
\end{center}
\end{table}
\setlength{\tabcolsep}{1.4pt}
\begin{myfigure*}[t]
    \centering
    \includegraphics[width=0.90\linewidth]{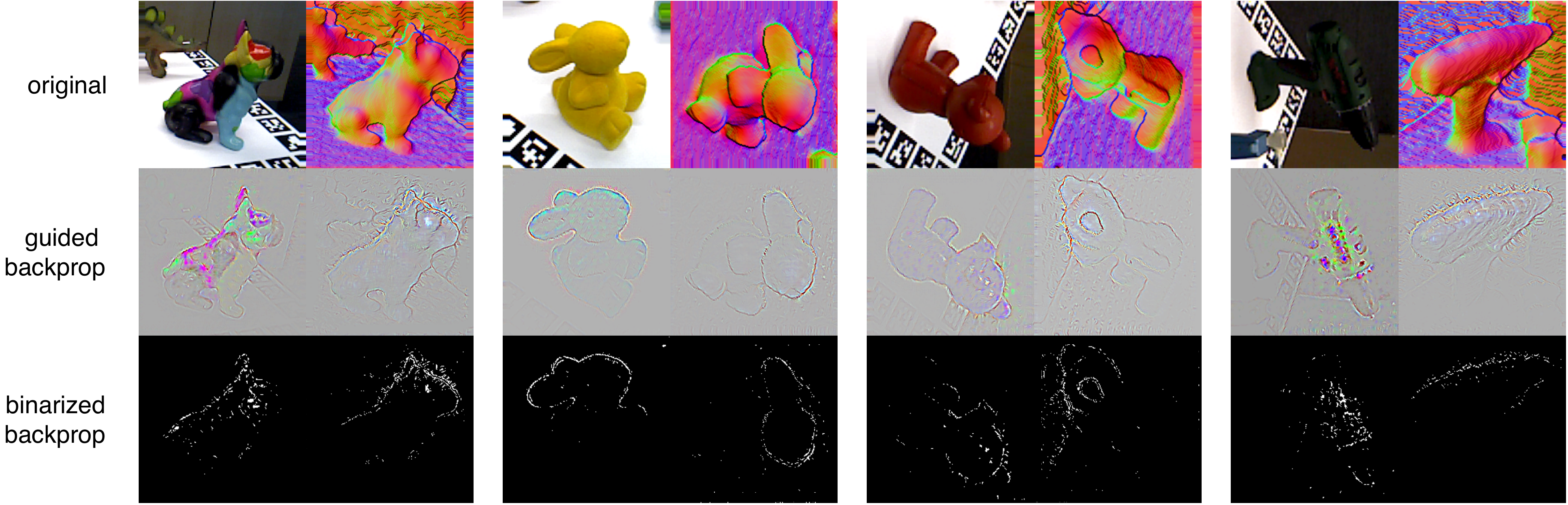}
    \caption{Visualization of the important pixels to predict the relative rotation. ``original" indicates the RGB-D input of the network; ``guided backprop"~\cite{springenberg2014striving} indicates the saliency map of the input based on the last layer of the feature extractors $E^c$ and $E^d$; ``binary backprop" is the binarized version of ``guided backprop" that highlights the peak values in white to facilitate visualization. The depth image is used with surfare normal colorization~\cite{aakerberg2017improving}.}
    \label{fig:guided_backprop}
\end{myfigure*}{}
\begin{figure}[t]
    \centering
    \includegraphics[width=1.0\linewidth]{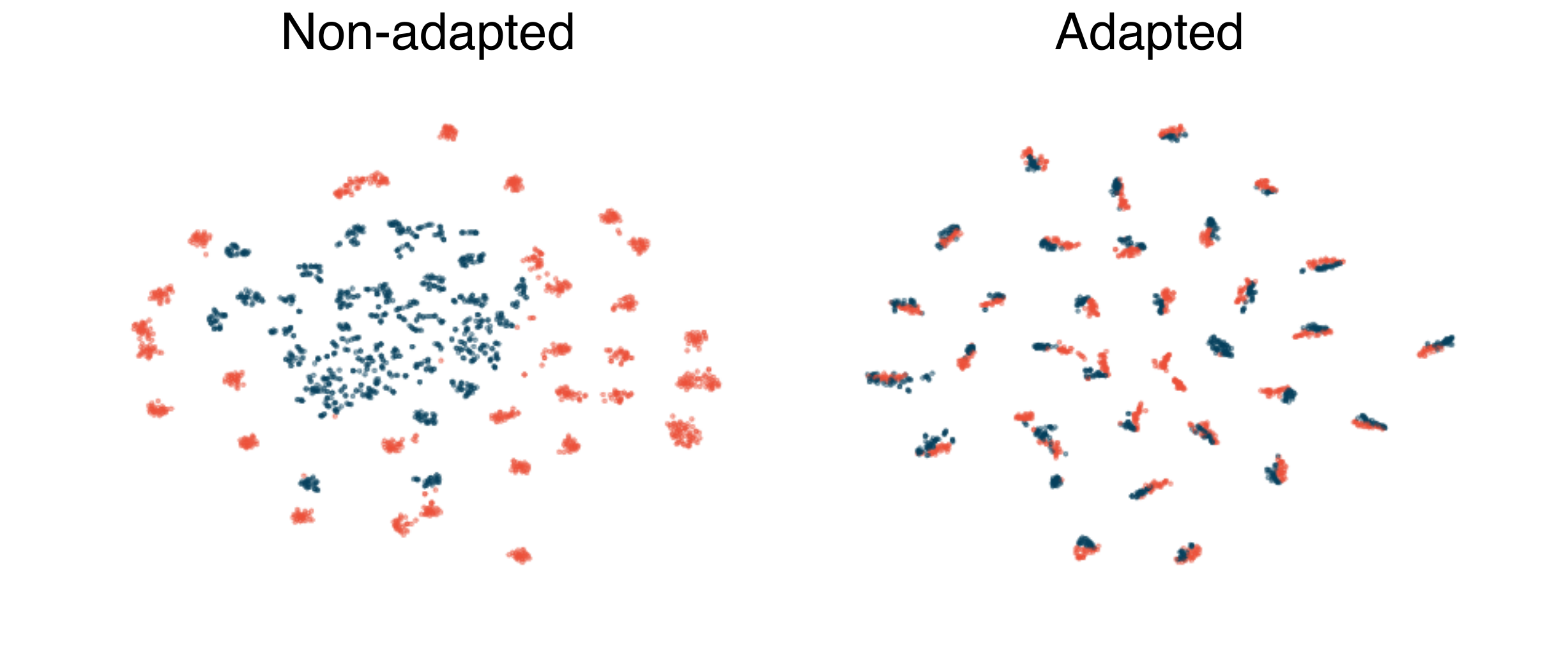}
    \caption{t-SNE~\cite{maaten2008visualizing} visualization of the HomebrewedDB~\cite{kaskman2019homebreweddb} features extracted from the last hidden layer of the main head $M$. Red dots: source samples; blue dots: target samples. When adapting the two domains with our method (right), the two distributions align much better compared to the non-adapted case (left).}
    \label{fig:tsne}
\end{figure}
\subsection{Results}
\label{subsec:results}

Table~\ref{tab:da} and~\ref{tab:ablation} present the quantitative results of RGB-D DA on the two benchmark datasets, synROD$\rightarrow$ROD and synHB$\rightarrow$realHB, while figure~\ref{fig:guided_backprop} provides qualitative insights into the functioning of our method. This empirical evaluation allows us to answer important research questions.

\textit{Are standard DA methods effective on multi-modal data?} Table~\ref{tab:da} shows that applying a standard DA method on RGB-D data is not always effective. For example, MMD and AFN perform worse when applied on the concatenation of RGB and depth features (RGB-D, RGB-D e2e) than when applied on the RGB features alone on synROD$\rightarrow$ROD. These results are due to the fact that the depth modality is far less informative than the RGB for object recognition when compared in isolation. Therefore, in the absence of an effective strategy to exploit both modalities, the RGB-D case can provide lower accuracy than the RGB alone. Comparing the two strategies to apply the baseline methods on multi-modal data (``RGB-D" and ``RGB-D e2e"), we noticed that no strategy clearly outperforms the other and the results are different depending on the method and the dataset used. It is also interesting to notice that AFN is not the best performing baseline on RGB-D data, despite being the considered the current state-of-the-art in DA. 

\textit{Is the relative rotation an effective pretext task to perform RGB-D DA?} Table~\ref{tab:da} shows that predicting the relative rotation between the RGB and depth image is indeed an effective DA strategy, significantly improving over ``Source only". This is also confirmed in figure~\ref{fig:tsne} where the t-SNE~\cite{maaten2008visualizing} visualization of the features of the main head $M$ show that our method effectively aligns the target and source distributions. More importantly, our method outperforms all considered baselines on both datasets. Compared to \textit{Rotation}, that is the most related to our method, we have $+3.35\%$ improvement on synROD$\rightarrow$ROD and $+2.82\%$ improvement on synHB$\rightarrow$realHB.

\textit{How are the different components of our method affecting the final performance?} We perform an ablation study to understand the impact of different components of our method on the overall performance. Following the example of~\cite{xu2019self-supervised}, we investigate what happens when we only use the target domain to solve the pretext task, instead of using both domains. Table~\ref{tab:ablation} shows that predicting the relative rotation of target samples is already sufficient to provide a significant improvement over ``Source only", but it is not as effective as using both domains. The network learns more informative features when solving the pretext task with both domains due to the higher diversity in the data. Finally, we evaluate the performance of our method when defining the pretext head $P$ with the same architecture as $M$. Table~\ref{tab:ablation} shows that this configuration leads to an average drop of $-1.64\%$ in accuracy. The results confirm that using convolutional layers instead of a pooling layer helps to better retain the spatial information necessary to predict the relative rotation.

\textit{What does the network learn to solve the relative rotation task?} 
Figure~\ref{fig:guided_backprop} shows the most relevant pixels to predict the relative rotation for a few example samples in realHB. More precisely, we use guided backpropagation~\cite{springenberg2014striving} to visualize which pixels of the RGB and depth input image maximally activate the last layer of the $E^c$ and $E^d$ to produce the correct prediction for the pretext task. First, we can notice that the most relevant pixels belong to the object, not the background or other elements in the image. This confirms that the network relies on the appearance of the object to make the prediction rather than learning ``trivial" shortcuts~\cite{doersch2015unsupervised}. Second, we can see that the network focuses on the same part of the object (e.g. the head of the bunny) in the RGB and depth image. This confirms that the prediction on the relative rotation is made by matching corresponding parts of the object in the two modalities. 

%% file: sections/conclusion.tex
In this work, we propose the first method tailored to tackle the challenging problem of RGB-D DA. Our approach consists of training a network to solve the self-supervised task of predicting the relative rotation between the RGB and depth image, in addition to the main object recognition task. To evaluate the performance of our method, we define two synthetic-to-real benchmarks for instance recognition and object categorization, using the existing HB and a newly collected dataset called synROD. We empirically demonstrate that our self-supervised task successfully reduces the domain shift and outperforms all considered baselines, indicating that exploiting the inter-modal relations is key to perform DA on RGB-D data. We hope that our work will ignite further research on the problem of DA for RGB-D and multi-modal data in general.

%% file: sections/ack.tex
This work has received funding from the European Union’s Horizon 2020 research and innovation program under the Marie Skłodowska-Curie grant agreement No. 676157, project ACROSSING, by the ERC grant 637076 - RoboExNovo.

%% file: root.bbl
\begin{thebibliography}{10}
\providecommand{\url}[1]{#1}
\csname url@samestyle\endcsname
\providecommand{\newblock}{\relax}
\providecommand{\bibinfo}[2]{#2}
\providecommand{\BIBentrySTDinterwordspacing}{\spaceskip=0pt\relax}
\providecommand{\BIBentryALTinterwordstretchfactor}{4}
\providecommand{\BIBentryALTinterwordspacing}{\spaceskip=\fontdimen2\font plus
\BIBentryALTinterwordstretchfactor\fontdimen3\font minus
  \fontdimen4\font\relax}
\providecommand{\BIBforeignlanguage}[2]{{%
\expandafter\ifx\csname l@#1\endcsname\relax
\typeout{** WARNING: IEEEtran.bst: No hyphenation pattern has been}%
\typeout{** loaded for the language `#1'. Using the pattern for}%
\typeout{** the default language instead.}%
\else
\language=\csname l@#1\endcsname
\fi
#2}}
\providecommand{\BIBdecl}{\relax}
\BIBdecl

\bibitem{krizhevsky2012imagenet}
A.~Krizhevsky, I.~Sutskever, and G.~Hinton, ``Imagenet classification with deep
  convolutional neural networks,'' in \emph{NeurIPS)}, 2012, pp. 1097--1105.

\bibitem{blender}
``Blender,'' \url{http://www.blender.org}, accessed: 2020-01-30.

\bibitem{aakerberg2017improving}
A.~Aakerberg, K.~Nasrollahi, and T.~Heder, ``Improving a deep learning based
  rgb-d object recognition model by ensemble learning,'' in \emph{IPTA}, 2017,
  pp. 1--6.

\bibitem{long2015learning}
M.~Long, Y.~Cao, J.~Wang, and M.~I. Jordan, ``Learning transferable features
  with deep adaptation networks,'' in \emph{ICML}, 2015.

\bibitem{ganin2016domain}
Y.~Ganin, E.~Ustinova, H.~Ajakan, P.~Germain, H.~Larochelle, F.~Laviolette,
  M.~Marchand, and V.~Lempitsky, ``Domain-adversarial training of neural
  networks,'' \emph{JMLR}, vol.~17, no.~1, pp. 2096--2030, 2016.

\bibitem{russo2018from}
P.~Russo, F.~M. Carlucci, T.~Tommasi, and B.~Caputo, ``From source to target
  and back: symmetric bi-directional adaptive gan,'' in \emph{CVPR}, 2018.

\bibitem{hoffman2017cycada}
J.~Hoffman, E.~Tzeng, T.~Park, J.-Y. Zhu, P.~Isola, K.~Saenko, A.~A. Efros, and
  T.~Darrell, ``Cycada: Cycle-consistent adversarial domain adaptation,'' in
  \emph{ICML}, 2017.

\bibitem{kaskman2019homebreweddb}
R.~Kaskman, S.~Zakharov, I.~Shugurov, and S.~Ilic, ``Homebreweddb: Rgb-d
  dataset for 6d pose estimation of 3d objects,'' in \emph{ICCV Workshops},
  2019, pp. 0--0.

\bibitem{lai2014unsupervised}
K.~Lai, L.~Bo, and D.~Fox, ``Unsupervised feature learning for {3D} scene
  labeling,'' in \emph{ICRA}, 2014, pp. 3050--3057.

\bibitem{sun2016return}
B.~Sun, J.~Feng, and K.~Saenko, ``Return of frustratingly easy domain
  adaptation,'' in \emph{AAAI}, 2016.

\bibitem{xu2019larger}
R.~Xu, G.~Li, J.~Yang, and L.~Lin, ``Larger norm more transferable: An adaptive
  feature norm approach for unsupervised domain adaptation,'' in \emph{ICCV},
  October 2019.

\bibitem{tzeng2017adversarial}
E.~Tzeng, J.~Hoffman, K.~Saenko, and T.~Darrell, ``Adversarial discriminative
  domain adaptation,'' in \emph{CVPR}, 2017.

\bibitem{ghifary2016deep}
M.~Ghifary, W.~B. Kleijn, M.~Zhang, D.~Balduzzi, and W.~Li, ``Deep
  reconstruction-classification networks for unsupervised domain adaptation,''
  in \emph{ECCV}, 2016.

\bibitem{bousmalis2016domain}
K.~Bousmalis, G.~Trigeorgis, N.~Silberman, D.~Krishnan, and D.~Erhan, ``{Domain
  Separation Networks},'' in \emph{NeurIPS}, 2016.

\bibitem{xu2019self-supervised}
J.~{Xu}, L.~{Xiao}, and A.~M. {López}, ``Self-supervised domain adaptation for
  computer vision tasks,'' \emph{IEEE Access}, vol.~7, pp. 156\,694--156\,706,
  2019.

\bibitem{carlucci2019domain}
F.~M. Carlucci, A.~D'Innocente, S.~Bucci, B.~Caputo, and T.~Tommasi, ``Domain
  generalization by solving jigsaw puzzles,'' in \emph{CVPR}, 2019, pp.
  2229--2238.

\bibitem{qi2018unified}
F.~Qi, X.~Yang, and C.~Xu, ``A unified framework for multimodal domain
  adaptation,'' in \emph{ACM Multimedia}, 2018, pp. 429--437.

\bibitem{spinello2012leveraging}
L.~Spinello and K.~O. Arras, ``Leveraging rgb-d data: Adaptive fusion and
  domain adaptation for object detection,'' in \emph{ICRA}.\hskip 1em plus
  0.5em minus 0.4em\relax IEEE, 2012, pp. 4469--4474.

\bibitem{hoffman2016cross-modal}
J.~Hoffman, S.~Gupta, J.~Leong, S.~Guadarrama, and T.~Darrell, ``Cross-modal
  adaptation for rgb-d detection,'' in \emph{ICRA}.\hskip 1em plus 0.5em minus
  0.4em\relax IEEE, 2016, pp. 5032--5039.

\bibitem{li2017domain}
X.~Li, M.~Fang, J.-J. Zhang, and J.~Wu, ``Domain adaptation from rgb-d to rgb
  images,'' \emph{Signal Processing}, vol. 131, pp. 27--35, 2017.

\bibitem{jing2019unsupervised}
W.~Jing and Z.~Kuangen, ``Unsupervised domain adaptation learning algorithm for
  rgb-d staircase recognition,'' \emph{arXiv preprint arXiv:1903.01212}, 2019.

\bibitem{doersch2015unsupervised}
C.~Doersch, A.~Gupta, and A.~A. Efros, ``Unsupervised visual representation
  learning by context prediction,'' in \emph{ICCV}, 2015, pp. 1422--1430.

\bibitem{noroozi2016unsupervised}
M.~Noroozi and P.~Favaro, ``Unsupervised learning of visual representations by
  solving jigsaw puzzles,'' in \emph{ECCV}.\hskip 1em plus 0.5em minus
  0.4em\relax Springer, 2016, pp. 69--84.

\bibitem{zhang2016colorful}
R.~Zhang, P.~Isola, and A.~A. Efros, ``Colorful image colorization,'' in
  \emph{ECCV}.\hskip 1em plus 0.5em minus 0.4em\relax Springer, 2016, pp.
  649--666.

\bibitem{pathak2016context}
D.~Pathak, P.~Krahenbuhl, J.~Donahue, T.~Darrell, and A.~A. Efros, ``Context
  encoders: Feature learning by inpainting,'' in \emph{CVPR}, 2016, pp.
  2536--2544.

\bibitem{gidaris2018unsupervised}
S.~Gidaris, P.~Singh, and N.~Komodakis, ``Unsupervised representation learning
  by predicting image rotations,'' \emph{arXiv preprint arXiv:1803.07728},
  2018.

\bibitem{golan2018deep}
I.~Golan and R.~El-Yaniv, ``Deep anomaly detection using geometric
  transformations,'' in \emph{NeurIPS}, 2018, pp. 9758--9769.

\bibitem{wu20153d}
Z.~Wu, S.~Song, A.~Khosla, F.~Yu, L.~Zhang, X.~Tang, and J.~Xiao, ``3d
  shapenets: A deep representation for volumetric shapes,'' in \emph{CVPR},
  2015, pp. 1912--1920.

\bibitem{yi2017large-scale}
L.~Yi, L.~Shao, M.~Savva, H.~Huang, Y.~Zhou, Q.~Wang, B.~Graham, M.~Engelcke,
  R.~Klokov, V.~Lempitsky \emph{et~al.}, ``Large-scale 3d shape reconstruction
  and segmentation from shapenet core55,'' \emph{arXiv preprint
  arXiv:1710.06104}, 2017.

\bibitem{loghmani2019recurrent}
M.~R. {Loghmani}, M.~{Planamente}, B.~{Caputo}, and M.~{Vincze}, ``Recurrent
  convolutional fusion for rgb-d object recognition,'' \emph{RA-L}, vol.~4,
  no.~3, pp. 2878--2885, July 2019.

\bibitem{carlucci2018deco}
F.~M. Carlucci, P.~Russo, and B.~Caputo, ``{(DE)$^2$CO}: Deep depth
  colorization,'' \emph{RA-L}, vol.~3, no.~3, pp. 2386--2396, 2018.

\bibitem{eitel2015multimodal}
A.~Eitel, T.~Springenberg, L.~S.~M. Riedmiller, and W.~Burgard, ``Multimodal
  deep learning for robust rgb-d object recognition,'' in \emph{IROS}, 2015,
  pp. 681--687.

\bibitem{lin2014microsoft}
T.-Y. Lin, M.~Maire, S.~Belongie, J.~Hays, P.~Perona, D.~Ramanan,
  P.~Doll{\'a}r, and C.~L. Zitnick, ``Microsoft coco: Common objects in
  context,'' in \emph{ECCV}.\hskip 1em plus 0.5em minus 0.4em\relax Springer,
  2014, pp. 740--755.

\bibitem{li2017deeper}
D.~Li, Y.~Yang, Y.-Z. Song, and T.~Hospedales, ``Deeper, broader and artier
  domain generalization,'' in \emph{ICCV}, 2017.

\bibitem{he2016deep}
K.~He, X.~Zhang, S.~Ren, and J.~Sun, ``Deep residual learning for image
  recognition,'' in \emph{CVPR}, 2016, pp. 770--778.

\bibitem{morerio2018minimal-entropy}
\BIBentryALTinterwordspacing
P.~Morerio, J.~Cavazza, and V.~Murino, ``Minimal-entropy correlation alignment
  for unsupervised deep domain adaptation,'' \emph{ICLR}, 2018. [Online].
  Available: \url{https://openreview.net/forum?id=rJWechg0Z}
\BIBentrySTDinterwordspacing

\bibitem{deng2009imagenet}
J.~Deng, W.~D.~R. Socher, L.~Li, K.~Li, and F.~Li, ``Imagenet: A large-scale
  hierarchical image database,'' in \emph{CVPR}, 2009, pp. 248--255.

\bibitem{springenberg2014striving}
J.~T. Springenberg, A.~Dosovitskiy, T.~Brox, and M.~Riedmiller, ``Striving for
  simplicity: The all convolutional net,'' \emph{arXiv preprint
  arXiv:1412.6806}, 2014.

\bibitem{maaten2008visualizing}
L.~v.~d. Maaten and G.~Hinton, ``Visualizing data using t-sne,'' \emph{JMLR},
  vol.~9, no. Nov, pp. 2579--2605, 2008.

\end{thebibliography}
